\documentclass[runningheads]{llncs}
\usepackage{graphicx}
\usepackage{bbding}
\usepackage[utf8]{inputenc}
\usepackage{amssymb}
\usepackage{graphicx}
\usepackage{multirow}
\usepackage{amsmath}% used for boldsymbol.
\usepackage{cite}

\pdfoutput=1

\graphicspath{{pic/}}

\begin{document}
\title{A Deep  Learning Method for Complex Human Activity Recognition Using Virtual Wearable Sensors}
\author{Fanyi Xiao\inst{1} \and
Ling Pei\inst{1(}\Envelope\inst{)}  \and
Lei Chu\inst{1} \and
Danping Zou\inst{1} \and
Wenxian Yu\inst{1} \and
Yifan Zhu\inst{1} \and
Tao Li\inst{1}}
\authorrunning{Fanyi Xiao et al.}
\institute{Shanghai Key Laboratory of Navigation and Location-based Services\\
School of Electrical Information and Electrical Engineering \\
Shanghai Jiao Tong University, Shanghai, China\\
\url{https://nls.sjtu.edu.cn/}}
\maketitle              % typeset the header of the contribution
\begin{abstract}
Sensor-based human activity recognition (HAR) is now a research hotspot in multiple application areas. With the rise of smart wearable devices equipped with inertial measurement units (IMUs), researchers begin to utilize IMU data for HAR. By employing machine learning algorithms, early IMU-based research for HAR can achieve accurate classification results on traditional classical HAR datasets, containing only simple and repetitive daily activities. However, these datasets rarely display a rich diversity of information in real-scene. In this paper, we propose a novel method based on deep learning for complex HAR in the real-scene. Specially, in the off-line training stage, the AMASS dataset, containing abundant human poses and virtual IMU data, is innovatively adopted for enhancing the variety and diversity. Moreover, a deep convolutional neural network with an unsupervised penalty is proposed to automatically extract the features of AMASS and improve the robustness. In the on-line testing stage, by leveraging advantages of the transfer learning, we obtain the final result by fine-tuning the partial neural network (optimizing the parameters in the fully-connected layers) using the real IMU data. The experimental results show that the proposed method can surprisingly converge in a few iterations and achieve an accuracy of 91.15\% on a real IMU dataset, demonstrating the efficiency and effectiveness of the proposed method.

\end{abstract}
%
%
%
% to the first part, namely introduction
\section{Introduction}

Human activity recognition (HAR) is a research hotspot in the field of computer vision and has broad application prospects in security monitoring, biological health, and other fields. Traditional recognition algorithms are mainly based on images or videos\cite{zhang2017review}. With the emergence of various wearable smart devices embedded with microsensors such as inertial measurement units (IMUs), these devices are highly used in daily life and play an indispensable role in emerging fields that strongly demand HAR such as virtual reality (VR). Therefore, it is a natural way to realize HAR based on wearable devices.

In recent years, HAR based on wearable devices has been conducted deep studies \cite{sousa2019human}, and there exist two general methods. Previous researches use traditional machine learning methods such as Support Vector Machine (SVM) and Random Forest (RF) to receive the recognition result \cite{pei2013human,pei2012using}. However, these methods need to to design features manually, calculate time and frequency domain features based on characteristics of the data. To reduce the computational consumption and compress input data, a further selection of features also needs to be conducted. Due to the longtime design and selection of manual features, it always costs lots using traditional methods of machine learning. With the development of deep learning in recent years, deep neural networks such as Convolutional Neural Network (CNN)\cite{ronao2016human} or Long Short-Term Memory networks (LSTM)\cite{chevalier2016lstms} have been widely used for HAR, finishing both feature extraction and activity classification.  

Almost all the above methods now can achieve excellent results on specific sensor-based HAR datasets. The widely used public datasets and their main characteristics are shown in Table~\ref{tab1}.

\begin{table}
\caption{Widely used public datasets and main characteristics (Acc=accelerometer, Gyro=gyroscope, Mag=Magnetometer, Temp=Temperature).}\label{tab1}
\begin{tabular}{c|c|c|c|c}
\hline
Datasets & Sampling Rate (Hz) & Sensors & Activities & Subjects\\
\hline
UCI HAR\cite{anguita2013public} & 50 & 2 (Acc, Gyro) & 6 & 30\\
% USC-HAD\cite{zhang2012usc} & 100 & 2 (Acc, Gyro) & 12 & 14\\
WISDM\cite{lockhart2011design} & 20 & 1 (Acc) & 7 & 36\\
WHARF\cite{bruno2015wearable} & 32 & 1 (Acc) & 12 & 17\\
PAMAP2\cite{reiss2012introducing} & 100 & 4 (Acc, Gyro, Mag, Temp) & 18 & 9\\
\hline
\end{tabular}
\end{table}

However, all these datasets have defects as follows:

\begin{itemize}
\item The most widely used datasets such as UCI HAR\cite{anguita2013public} contain only simple daily activities, for example, walking, running or jumping, while human behaves much more complex in real life.
\item Subjects involved in data collection are always limited, and the same activity tends to be performed similarly, for instance, walking may only include walking at normal speed. However, the same activity can be performed in different styles and may vary with different humans in the real world.
\item During data collection, most datasets use only a single IMU, which makes them unsuitable for recognizing more elaborate activities such as stretching arms or stretching legs. Though other datasets use more than two IMUs, the increase in IMUs also leads to the intrusion to subjects.
\end{itemize}

To solve problems above, this paper innovatively adopts a pose reconstruction dataset AMASS\cite{mahmood2019amass}, which is a large collection of motion capture (Mocap) datasets, for HAR. The adoption of this dataset has the following advantages:

\begin{enumerate}
\item AMASS contains rich motion types. It includes complex activities such as house cleaning in addition to simple daily activities, making this dataset closer to real life.
\item The containing of multiple mocap datasets in AMASS leads to both richer characteristics in activities and an increase in the number of involved subjects, which is more than 300. 
\item Inspired by \cite{huang2019deep}, where virtual IMU data are innovatively used in pose reconstruction, we similarly use virtual IMU for HAR, which greatly reduces the cost of collecting real datasets.
\end{enumerate}

The main contributions of this paper are as follows:
\begin{itemize}
\item[-] Adopt a novel pose reconstruction dataset AMASS for HAR and use virtual IMU data in this dataset.
\item[-] Use a realistic dataset to fine-tune the model for further reducing the gap between real and virtual data.
\item[-] Propose a CNN framework combined with an unsupervised penalty for HAR. 
\end{itemize}

Experimental results show that test result on the realistic dataset is 91.15\% after fine-tuning, which demonstrates the feasibility of applying pose recognition datasets and using virtual IMU data for HAR. 
%
%
%
% to next part, namely, dataset preprocessing
\section{Dataset preprocessing based on the SMPL model}

One major work of this paper is the processing of AMASS, making it suitable for HAR. Since the IMU data in AMASS is virtual, this paper further processes the DIP dataset proposed in \cite{huang2019deep}, which contains real IMU data that can be used to reduce the gap between virtual data and real data.  

\subsection{SMPL model}

SMPL \cite{loper2015smpl} is a parameterized model of the 3D human body, totally including $N = 6890$ vertices, $K = 23$ joints. Input parameters of this model are shape parameters $ \boldsymbol{\beta} $, which takes 10 values controlling the shape change of the human body, and pose parameters $\boldsymbol{\theta}$ that takes 72 values which define the relative angles of 24 joints (including the root joint) of the human body:

\begin{equation}
    M(\boldsymbol{\beta, \theta}) = W(T_P(\boldsymbol{\beta,\theta}),J(\boldsymbol{\beta}),\boldsymbol{\theta},\boldsymbol{W}),
\end{equation}

\begin{equation}    
    T_P(\boldsymbol{\beta, \theta}) = \boldsymbol{T}+B_s(\boldsymbol{\beta})+B_p(\boldsymbol{\theta}),
\end{equation}
where $T$ defines a template mesh, to which pose-dependent deformations $B_s(\boldsymbol{\beta})$ and shape-dependent deformations $B_p(\boldsymbol{\theta})$ are added. Based on the rotation around the predicted joint locations $J(\boldsymbol{\beta})$ with smoothing defined by the blend weight matrix $\boldsymbol{W}$, the resulting mesh is then posed using a standard linear skinning function (LBS). 

Using this model, AMASS converts the motion poses of several classical motion capture datasets such as Biomotion \cite{troje2002decomposing}, from a skeletal form to a more realistic 3D skin model, while the pose parameters are given as a rotation matrix.

\subsection{Virtual data generation}

Though AMASS contains the input parameters of the SMPL model, it does not contain IMU data as original mocap datasets do not provide IMU data. To use AMASS for sensor-based pose reconstruction, \cite{huang2019deep} confirms the feasibility of synthesizing IMU data and generating corresponding SMPL parameters based on the input of different models.

Based on the rich information provided by AMASS, virtual acceleration data and orientation readings in the rotation matrix can be generated by placing virtual sensors on the SMPL mesh surface. Orientation readings are directly obtained using forward kinematics, while virtual accelerations are calculated via finite differences \cite{huang2019deep}. The virtual acceleration for time $t$ is defined as:
\begin{equation}
a_t=\frac{p_{t-1} + p_{t+1} - 2\cdot p_t}{dt^2},
\end{equation}
where $p_t$ is the position of a virtual IMU for time $t$, and $d_t$ is the time interval between two consecutive frames.

\subsection{Labeling and filtering with SMPL model}

Since AMASS contains over 11000 motions, it is necessary to classify these motions into different activities and make true labels. Further, a single motion file in AMASS may consist of several activities, so it is also essential to filter out some motions that affect the balance of the dataset. Activity labeling and data filtering are mainly achieved through three steps. 

\subsubsection{Posture-based labeling}

We first classify the whole motions in AMASS into 12 categories based on the superficial descriptions of motions in most classical mocap datasets included in AMASS. Two types of motions are directly removed in this procedure. The first type is motions with little relevance to human daily activities, such as boxing and other martial motions described in Biomotion \cite{troje2002decomposing}. The second type refers to some frequently converted motions(e.g. quick transitions between walking, stopping and running). Since the motion duration is generally short in AMASS, frequent motion transitions may conflict with the subsequent sliding window length settings, therefore such motions are also excluded.

\subsubsection{Acceleration-based filtering}

\begin{figure}
\centering
\begin{minipage}[b]{0.45\linewidth}
    \includegraphics[width=0.98\linewidth]{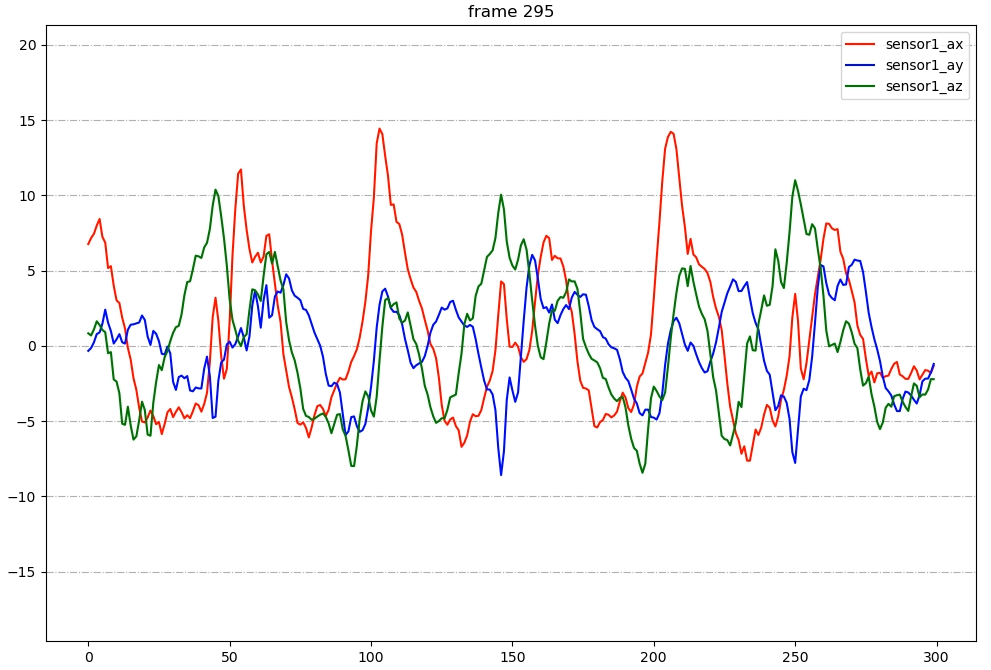}
    \caption{Wrist accelerations for typical walking movement}
    \label{fig1}
\end{minipage}
\quad
\begin{minipage}[b]{0.45\linewidth}
    \includegraphics[width=0.98\linewidth]{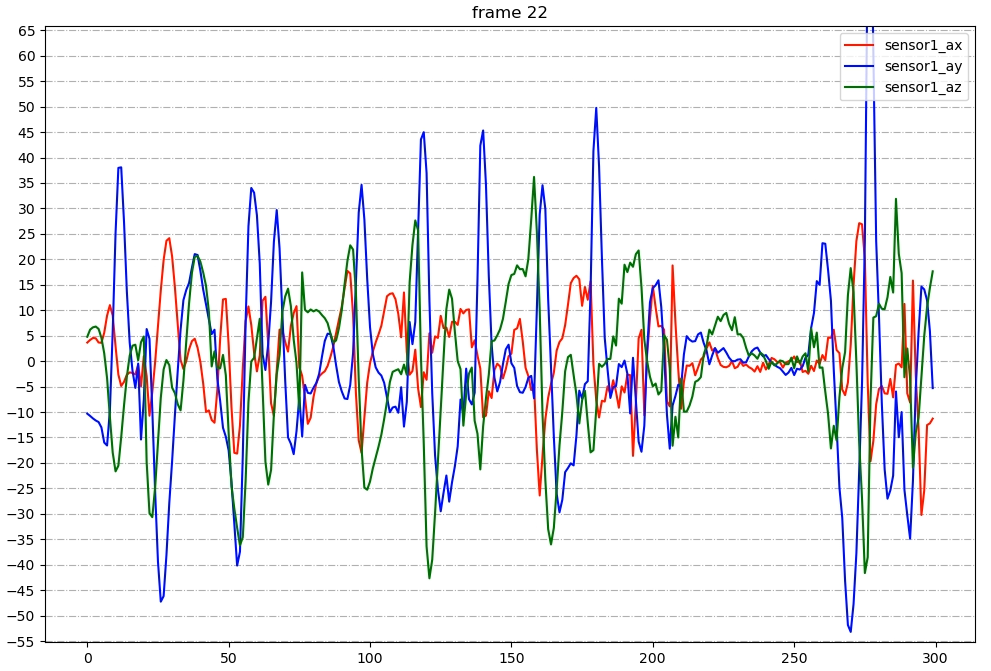}
    \caption{Wrist accelerations for typical running movement}
    \label{fig2}
\end{minipage}
\end{figure}

A simple classification of the dataset is implemented in the previous section, while some data are further filtered based on accelerations. Using the accelerations obtained via the sensor on the wrist, a dynamic graph of the acceleration over time can be created. The accelerations at the left wrist for typical walking and running movements are shown in Fig.~\ref{fig1} and Fig.~\ref{fig2}.

As can be seen from the comparison of Fig.~\ref{fig1} and Fig.~\ref{fig2}, different activities often differ in acceleration characteristics. Therefore, data is further cleaned based on the differences in the characteristics of acceleration (e.g. peaks, variances, etc.).

\subsubsection{Data cleaning with SMPL model}

For some activities whose acceleration characteristics are not obvious, such as the stretching of the arms, it almost fails using acceleration features to clean the dataset. However, since AMASS provides SMPL pose parameters in the form of the rotation matrix, it becomes feasible to filter this type of activity adopting visualization with the SMPL model.

\begin{figure}[htpb]
\centering
\begin{minipage}[b]{0.45\linewidth}
    \centering
    \includegraphics[height=0.7\linewidth]{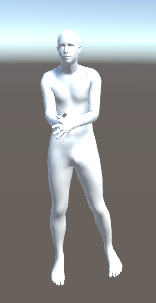}
    \caption{Visualization for clapping movement}
    \label{fig3}
\end{minipage}
\quad
\begin{minipage}[b]{0.45\linewidth}
    \centering
    \includegraphics[height=0.7\linewidth]{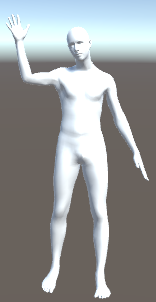}
    \caption{Visualization for waving movement}
    \label{fig4}
\end{minipage}
\end{figure}

After using Unity to build the SMPL model, the motions can be visualized by passing in different SMPL pose parameters. Clapping motion and motion of waving arms are shown in Fig.~\ref{fig3} and Fig.~\ref{fig4} respectively. After visualization of such data, mislabeled motions can be successfully deleted. 

However, preprocessed AMASS still suffers the problem of extremely unbalanced activities after processing above, which is mainly caused by unbalanced motions in the original AMASS. To alleviate this problem, interpolation up-sampling is adopted in this paper.

\section{Deep learning algorithm and fine-tuning}
\subsection{Proposed method}

The proposed method includes two stages: the off-line training stage and the on-line testing stage. At the first stage, we firstly employ the AMASS dataset, containing abundant human poses, to enhance the variety and diversity of the real data.

Motivated by the pioneer works \cite{Erhan2010Why,  Lecun2015Deep},  a deep convolutional neural network ($U$ convolutional layers and $S$ fully-connected layers) with an unsupervised penalty ($U$ deconvolutional layers)  is proposed to automatically extract the features of AMASS. Specially, given $p$-th batch IMU data ${{\bf{X}}^{\left( p \right)}}$ and the related labels ${{\bf{Z}}^{\left( p \right)}}$, the proposed method tries to update the neural network parameters $ \Theta  = {\Theta _0} \cup {\Theta _1}{\rm{ = }}{\left\{ {{{\bf{W}}_k}} \right\}_{k = 1,2, \cdots ,2U + S}} $ by minimizing
%$\Theta = {{\Theta _0} \cup \Theta_1 }$ by minimizing

\begin{equation}
\label{eq32}
\begin{array}{*{20}{c}}
{\mathop {{\rm{argmin}}}\limits_\Theta  }&{\underbrace {{\cal L}_{0}\left( {{\Theta _0}} \right)}_{supervised} + \lambda \underbrace {{{\cal L}_{1}}\left( {{\Theta _1}} \right)}_{unsupervised \ penalty}}
\end{array},
\end{equation}

where   \[ {\cal L}_{0}\left( \Theta_0  \right) =  \left\| {{{\bf{Z}}^{\left( p \right)}} - {\varphi _S}\left( {{{\bf{W}}_S}{{{\bf{\tilde X}}}^{\left( p \right)}}} \right)} \right\|_2^2, \] 
\[{{{\bf{\tilde X}}}^{\left( p \right)}} = {\varphi _U}\left( {{{\bf{W}}_U} \cdots {\varphi _1}\left( {{{\bf{W}}_1}{{\bf{X}}^{\left( p \right)}}} \right)} \right),\]
\[{{\cal L}_1}\left( {{\Theta _1}} \right) = \left\| {{{\bf{X}}^{\left( p \right)}} - {\varphi _{2U}}\left( {{{\bf{W}}_{2U}} \cdots {\varphi _{U + 1}}\left( {{{\bf{W}}_1}{{{\bf{\tilde X}}}^{\left( p \right)}}} \right)} \right)} \right\|_2^2,\]
$\varphi$ is the activation function of the $i$-th layer, and $\lambda$ is the penalty parameter that balances $\mathcal{L}\left(\Theta_0\right)$ and $\mathcal{L}_{1}\left(\Theta_{1}\right)$. We use an unsupervised penalty to promote the generalization of the proposed method by considering:
\begin{itemize}
	\item In our case, by optimizing $\mathcal{L}_{0}\left(\Theta_{0}\right)$, we try to represent ${{\bf{X}}^{\left( p \right)}}$ of high-dimension by the  latent layer of low-dimension (${{{\bf{\tilde X}}}^{\left( p \right)}}$). Such an operation, considering the low dimensionality of the IMU data, is helpful for the key feature extraction.   
	\item The unsupervised penalty in \eqref{eq32} itself is a denoising autoencoder \cite{Lecun2015Deep} that can help denoising the AMASS dataset, enhancing the robustness of the proposed method. 
	\item It has been shown in previous studies that learning multi-task (i.e., $\mathcal{L}_{0}\left(\Theta_{0}\right)$ and $\mathcal{L}_{1}\left(\Theta_{0}\right)$) jointly can improve the generalization error bounds \cite{chu2019lemo, Maurer2013Excess}. 
\end{itemize}

\subsection{Fine-tuning with real IMU}

Since IMU data in AMASS is virtually generated via the SMPL model and virtual sensors, while the IMU data in the real world tends to be affected by environmental noise, electromagnetic waves, etc. Therefore, certain differences exist between virtual and real data. To eliminate the gap, this paper uses the DIP dataset with real IMU data provided in \cite{huang2019deep} for fine-tuning. Data processing of DIP is similar to AMASS, except the fact that DIP only contains 5 activities, namely “computer works”, “walking”, “jumping”, “stretching arms” and “stretching legs”. Meanwhile, DIP has rather balanced activity categories, therefore up-sampling is not performed on DIP. 

Following the off-line training stage in Section 3.1, at the on-line testing stage, by leveraging advantages of the transfer learning, we obtain the final result by fine-tuning the parameters in the fully-connected layers with the real IMU data.  

\section{Test Verification}

This paper innovatively adopts a pose reconstruction dataset AMASS with virtual IMU data for HAR and proposes a new CNN framework with an unsupervised penalty. We design several comparative experiments, to prove the feasibility of using pose reconstruction dataset for HAR. 

To further verify the rationality of the method proposed in this paper, both classical machine learning algorithms and deep learning algorithms are tested on AMASS and DIP. Taking the sequence length in AMASS into consideration, this paper finally adopts RF and DeepConvLSTM\cite{ordonez2016deep} algorithms for comparisons. For RF we directly input the processed data for classification, while we adopt the original DeepConvLSTM architecture for comparison.

\subsection{Experimental design}

Three groups of comparative experiments based on different datasets are designed. Experiment 1 conducts training and testing on AMASS, using all three algorithms. The ratio of the training set to the test set is 7:3. Experiment 2 conducts training and testing on DIP and adopts all three algorithms similar to experiment 1. Experiment 3 is trained on the AMASS training set, fine-tuned on the DIP training set and finally tested on the DIP test set. Only our proposed method and DeepConvLSTM are involved in experiment 3.  

Considering that some activities cannot be identified using only one IMU, three IMUs located at the left wrist, the right thigh, and the head are selected in this paper. The total input data have features in 36 dimensions, including three-axis acceleration and rotation matrix. Since the sampling rates of AMASS and DIP are both 60Hz, a sliding window with 60 frames (i.e. 1 second) length is selected, while the degree of overlapping is set as 50\%.

\subsection{Evaluation criteria}

Commonly used evaluation criteria in HAR are accuracy, recall, F1-score and Area Under the Curve (AUC), among which accuracy and F1-score are most commonly used. Therefore, we also adopt accuracy and F1-score as the performance measures:
\begin{equation}
Accuracy=\frac{\sum_{cn=1}^{CN}TP_{cn}+\sum_{cn=1}^{CN}TN_{cn}}{\sum_{cn=1}^{CN}TP_{cn}+\sum_{cn=1}^{CN}TN_{cn}+\sum_{cn=1}^{CN}FP_{cn}+\sum_{cn=1}^{CN}FN_{cn}},
\end{equation}
\begin{equation}
F1-score=\frac{2\sum_{cn=1}^{CN}TP_{cn}}{2\sum_{cn=1}^{CN}TP_{cn}+\sum_{cn=1}^{CN}FP_{cn}+\sum_{cn=1}^{CN}FN_{cn}},
\end{equation}
where $CN$ denotes the class number. Variables $TP_{cn}$, $FP_{cn}$, $TN_{cn}$, $FN_{cn}$ are the true positives, false positives, true negatives and false negatives of the class $cn$ , respectively.

\subsection{Experimental results and analysis}

Table~\ref{tab2} illustrates all results in three experiments. From the results on the AMASS dataset in Table~\ref{tab2}, we can see that all three algorithms can achieve accuracy over 70\%, despite the fact that IMU data in AMASS is virtual and the containing of complex activities composed of several motions. Results on the DIP dataset in Table~\ref{tab2} corresponds to the results of experimen2, comparing three algorithms on a realistic IMU dataset DIP. We can see that the proposed method outperforms DeepConvLSTM and RF on both AMASS and DIP, which strongly illustrates the rationality of the deep learning algorithm proposed in this paper. 

\begin{table}
    \centering
    \caption{Experimental results}
    \label{tab2}
    \begin{tabular}{c|c|c|c|c|c|c}
        \hline
        \multirow{3}{*}{Dataset} & 
            \multicolumn{6}{c}{Methods \& results} \\
            \cline{2-7}
            & \multicolumn{2}{c|}{Proposed method}
            & \multicolumn{2}{c|}{DeepConvLSTM}
            & \multicolumn{2}{c}{RF} \\
            \cline{2-7}
            & Acc & F1-score & Acc & F1-score
            & Acc & F1-score \\
        \hline
        AMASS & \textbf{87.46\%} & \textbf{86.50\%}
        & 73.03\% & 72.43\% & 75.01\% & 70.00\% \\
        DIP & \textbf{89.08\%} & \textbf{89.16\%}
        & 78.33\% & 79.31\% & 77.25\% & 75.96\% \\
        AMASS \& DIP & \textbf{91.15\%} & \textbf{91.21\%}
        & 84.80\% & 85.12\% & \verb|\| & \verb|\| \\
    \hline
    \end{tabular}
\end{table}

Notice that the classification result on DIP is not as good as the classification result of DeepConvLSTM and RF on classical HAR datasets. The main reason is that although DIP only contains 5 activities, similar to AMASS, each activity may be composed of a variety of motions, such as activity stretching legs which includes two motions, leg raising, and stepping. Activities with multiple motions greatly increase the difficulty of classification.

To confirm gaps between virtual IMU data and real IMU data, we additionally use the proposed network trained on AMASS to finish the classification task on DIP, an unsurprising result of accuracy less than 50\% is obtained. While the network trained based on AMASS and fine-tuned on the DIP training set achieves the best performance on the DIP test set, both for the proposed method and DeepConvLSTM. The results confirm that fine-tuning indeed eliminates the gap between the virtual IMU and the real IMU to some extent.

\begin{figure}[htp]
\centering
\begin{minipage}[b]{0.48\linewidth}
    \includegraphics[width=0.98\linewidth]{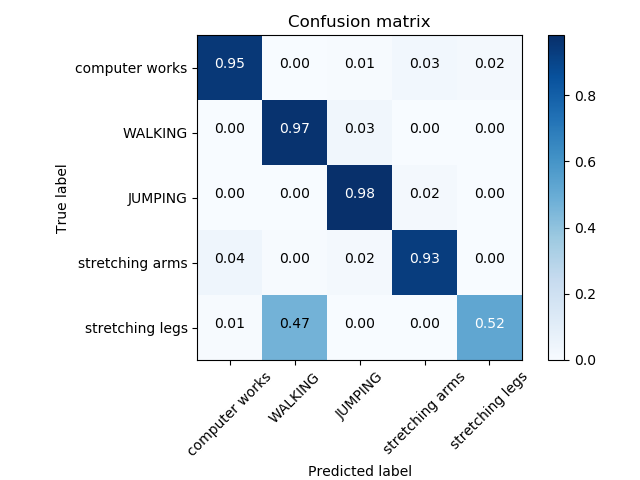}
    \caption{Confusion matrix of DIP in experiment 2}
    \label{fig5}
\end{minipage}
\quad
\begin{minipage}[b]{0.48\linewidth}
    \includegraphics[width=0.98\linewidth]{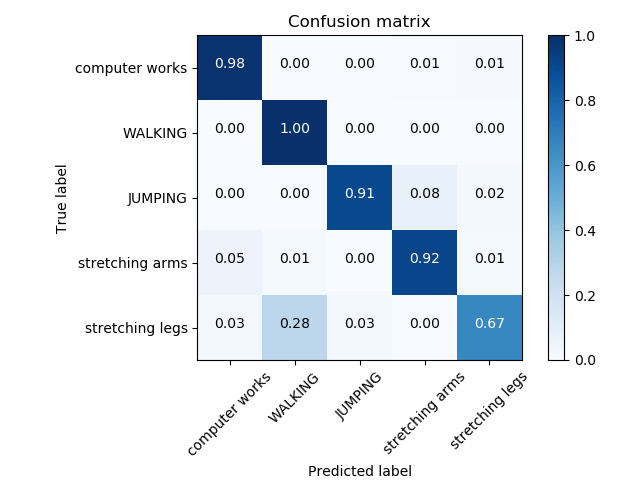}
    \caption{Confusion matrix of DIP in experiment3}
    \label{fig6}
\end{minipage}
\end{figure}

We also show the confusion matrix figures of the proposed method in experiment 2 experiment 3. As Fig.~\ref{fig5} and Fig.~\ref{fig6} show, fine-tuning effectively improves the classification results of some categories in DIP, which is mainly caused by richer motions in AMASS that make it easier to distinguish some confusing activities. Another interesting thing to be noticed is that fine-tuning can achieve rather excellent results within 20 epochs. This also provides a way for future research, that is, training on large-scale virtual IMU datasets, only need for a small scale of datasets with real IMU data for fine-tuning, which will reduce the cost of collecting real data.

\section{Conclusion}

This paper innovatively adopts a pose reconstruction dataset AMASS for HAR for the problem of simple daily activities and limited subjects in classical datasets. At the same time, a pose reconstruction dataset DIP with real IMU data is used for fine-tuning, to reduce the gap between virtual IMU data and real IMU data. Future work can focus on the most suitable IMU configurations through more detailed experiments.

\subsubsection{Acknowledgment}
This work was supported by the National Nature Science Foundation of China (NSFC) under Grant 61873163, Equipment Pre-Research Field Foundation under Grant 61405180205, Grant 61405180104.

\bibliographystyle{abbrv}
%\bibliography{reference1}

\end{document}